\documentclass[10pt,twocolumn,letterpaper]{article}

\usepackage{cvpr}              % To produce the CAMERA-READY version
\usepackage{graphicx}
\usepackage{amsmath}
\usepackage{amssymb}
\usepackage{booktabs}
\makeatletter
\@namedef{ver@everyshi.sty}{}
\makeatother
\usepackage{tikz}

\usepackage[pagebackref,breaklinks,colorlinks]{hyperref}

\usepackage[capitalize]{cleveref}
\crefname{section}{Sec.}{Secs.}
\Crefname{section}{Section}{Sections}
\Crefname{table}{Table}{Tables}
\crefname{table}{Tab.}{Tabs.}

\usepackage{booktabs}
\usepackage{multicol}
\usepackage{multirow}
\usepackage{color}
\usepackage{caption}
\usepackage{hhline}
\usepackage{pifont}
\usepackage{threeparttable}
\usepackage{makecell}
\usepackage{algorithm} 
\usepackage{listings}
\usepackage{amsfonts,amssymb}
\usepackage{graphicx}
\usepackage{lipsum}
\newcommand\blfootnote[1]{%
  \begingroup
  \renewcommand\thefootnote{}\footnote{#1}%
  \addtocounter{footnote}{-1}%
  \endgroup
}
\usepackage[accsupp]{axessibility}
\usepackage{colortbl}

\newcommand{\tabincell}[2]{\begin{tabular}{@{}#1@{}}#2\end{tabular}}

\newlength\savedwidth
\newcommand\whline{\noalign{\global\savedwidth\arrayrulewidth\global\arrayrulewidth 0.8pt}\hline\noalign{\global\arrayrulewidth\savedwidth}}
\usepackage{verbatim}

\definecolor{resnet}{rgb}{0.9607843137254902, 0.8705882352941177, 0.7019607843137254}
\definecolor{regnety}{rgb}{0.9764705882352941 , 0.792156862745098 , 0.1411764705882353}
\definecolor{vit}{rgb}{0.9019607843137255 , 0.403921568627451 , 0.403921568627451}
\definecolor{deit}{rgb}{1.0 , 0.7137254901960784 , 0.7568627450980392}
\definecolor{pvt}{rgb}{0.7686274509803922 , 0.27058823529411763 , 0.4117647058823529}
\definecolor{mlpmixer}{rgb}{0.0 , 0.5803921568627451 , 0.19607843137254902}
\definecolor{resmlp}{rgb}{0.25098039215686274 , 0.8784313725490196 , 0.8156862745098039}
\definecolor{swinmixer}{rgb}{0.2196078431372549 , 0.403921568627451 , 0.8392156862745098}
\definecolor{gmlp}{rgb}{0.0 , 0.7215686274509804 , 0.5803921568627451}
\definecolor{poolformer}{rgb}{0.8666666666666667 , 0.6274509803921569 , 0.8666666666666667}
\definecolor{resnetimproved}{rgb}{1.0 , 0.54902 , 0.0}

\newcommand{\resnetdot}{\raisebox{0.5pt}{\tikz\fill[resnet] (0,0) (0ex,-1ex)--(-0.866ex,0.5ex)--(0.866ex, 0.5ex)--cycle;}}

\newcommand{\deitdot}{\raisebox{0.5pt}{\tikz\fill[deit] (0,0) (0ex,1ex)--(-0.866ex,-0.5ex)--(0.866ex, -0.5ex)--cycle;}}
\newcommand{\pvtdot}{\raisebox{0.5pt}{\tikz\fill[pvt] (0,0) (0ex,1ex)--(-0.866ex,-0.5ex)--(0.866ex, -0.5ex)--cycle;}}
\newcommand{\mlpmixerdot}{\raisebox{0.5pt}{\tikz\fill[mlpmixer] (0,0) (1ex,0ex)--(-0.5ex,0.866ex)--(-0.5ex, -0.866ex)--cycle;}}
\newcommand{\resmlp}{\raisebox{0.5pt}{\tikz\fill[resmlp] (0,0) (1ex,0ex)--(-0.5ex,0.866ex)--(-0.5ex, -0.866ex)--cycle;}}
\newcommand{\swinmixer}{\raisebox{0.5pt}{\tikz\fill[swinmixer] (0,0) (1ex,0ex)--(-0.5ex,0.866ex)--(-0.5ex, -0.866ex)--cycle;}}

\newcommand{\poolformer}{\raisebox{0.5pt}{\tikz\fill[poolformer] (0,0) circle (.8ex);}}

\newcommand{\vect}[1]{\boldsymbol{\mathbf{#1}}}
\def\eg{\emph{e.g}\onedot} 
\def\ie{\emph{i.e}\onedot} 
 
\def\etc{\emph{etc}\onedot}

\newcommand{\gr}{\rowcolor[gray]{.95}}
\definecolor{throughput}{rgb}{0.0 , 0.5564705882352941 , 0.031372549019607}
\newcommand{\hlg}[1]{\textcolor{throughput}{#1}}

\newcommand{\better}[1]{\hlg{$\uparrow\,$#1}}

\usepackage[normalem]{ulem}
\newcommand\delete{\bgroup\markoverwith{\textcolor{red}{\rule[0.5ex]{2pt}{1.0pt}}}\ULon}

\usepackage{enumitem}
\def\cvprPaperID{2608} % *** Enter the CVPR Paper ID here
\def\confName{CVPR}
\def\confYear{2023}

\begin{document}

\title{RIFormer: Keep Your Vision Backbone Effective But Removing Token Mixer}

\author{
Jiahao Wang\textsuperscript{1,2}
\quad Songyang Zhang\textsuperscript{1}$^{\ast}$
\quad Yong Liu\textsuperscript{3}
\quad Taiqiang Wu\textsuperscript{3}
\quad Yujiu Yang\textsuperscript{3}
\\
\quad Xihui Liu\textsuperscript{2}
\quad Kai Chen\textsuperscript{1}$^{\ast}$ 
\quad Ping Luo\textsuperscript{2}
\quad Dahua Lin\textsuperscript{1}
\\
\textsuperscript{1}{Shanghai AI Laboratory}  \quad
\textsuperscript{2}{The University of HongKong} \\ \quad 
\textsuperscript{3}{Tsinghua Shenzhen International Graduate School} \\
\tt\small wang-jh19@tsinghua.org.cn \quad 
\{zhangsongyang,chenkai,lindahua\}@pjlab.org.cn \\
\tt\small yang.yujiu@sz.tsinghua.edu.cn \quad xihuiliu@eee.hku.hk \quad pluo@cs.hku.hk \\
}

\maketitle

%%%%%%%%% ABSTRACT
\begin{abstract}
This paper studies how to keep a vision backbone effective while removing token mixers in its basic building blocks. Token mixers, as self-attention for vision transformers (ViTs), are intended to perform information communication between different spatial tokens but suffer from considerable computational cost and latency. However, directly removing them will lead to an incomplete model structure prior, and thus brings a significant accuracy drop. To this end, we first develop an \underline{R}ep\underline{I}dentity\underline{Former} base on the re-parameterizing idea, to study the token mixer free model architecture. And we then explore the improved learning paradigm to break the limitation of simple token mixer free backbone, and summarize the empirical practice into 5 guidelines. Equipped with the proposed optimization strategy, we are able to build an extremely simple vision backbone with encouraging performance, while enjoying the high efficiency during inference. Extensive experiments and ablative analysis also demonstrate that the inductive bias of network architecture, can be incorporated into simple network structure with appropriate optimization strategy. We hope this work can serve as a starting point for the exploration of optimization-driven efficient network design. Project page: \url{https://techmonsterwang.github.io/RIFormer/}.

\small\blfootnote{\noindent$^{\ast}$ Corresponding author.}
\end{abstract}
%%%%%%%%% BODY TEXT
\section{Introduction}
\label{sec:intro}

The monumental advance in computer vision in the past few years was partly brought by the revolution of vision backbones, including \textit{convolutional neural networks (ConvNets)}~\cite{liu2022convnet, he2016deep, ding2021repvgg, rao2022hornet} and \textit{vision transformers (ViTs)}~\cite{dosovitskiy2020image, deit}. Both of them have particular modules in their basic building blocks that aggregate information between different spatial locations, which are called \textit{token mixer}~\cite{yu2022metaformer}, such as self-attention for ViTs. Although the effectiveness of token mixer has been demonstrated on many vision tasks~\cite{liu2021swin, detr, xie2021segformer, ipt}, its computational complexity typically takes up a significant portion of the network. In practice, heavy token mixers make the vision backbone limited especially on the edge-side devices due to the issue of speed and computation cost.

\begin{figure}[t]
	\centering
	\includegraphics[width=1.0\linewidth]{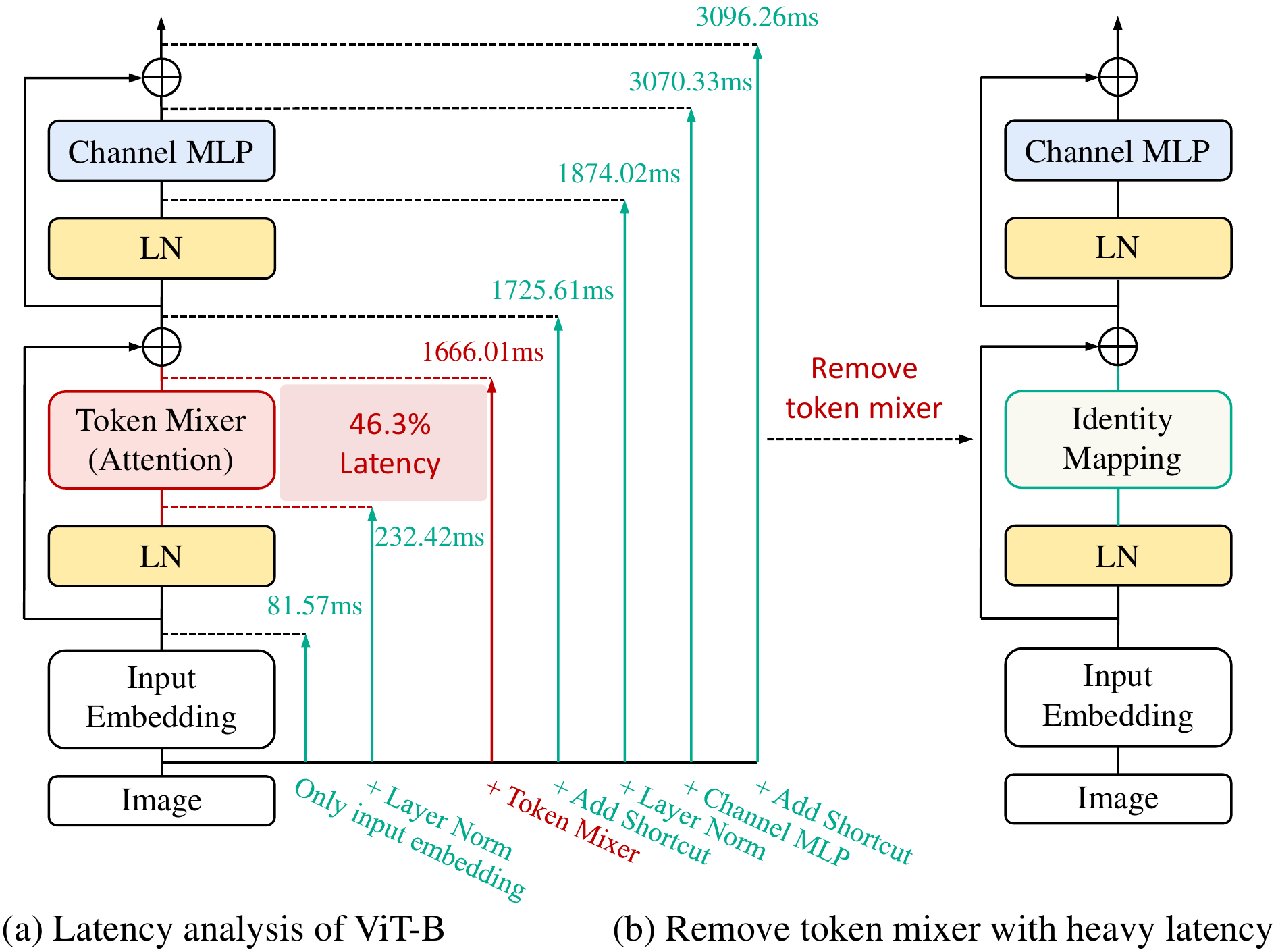}
    \vspace{-0.2in}
	\caption{Latency analysis of different components in  ViT-Base~\cite{dosovitskiy2020image}. (a) For token mixer (self-attention), the latency occupies about 46.3\% of the backbone. (b) Our motivation was to remove the token mixer while striving to keep the performance.}
	\label{fig:latency}
    \vspace{-0.2in}
\end{figure}

There have been several attempts in the literature to investigate efficient token mixers for slimming vision backbones\cite{rao2021global, yu2022metaformer,rao2021dynamicvit}. Although those works have already achieve competitive performance with light-weight design, they do retain the token mixers, which brings non-negligible increase in latency, as illustrated in Fig.~\ref{fig:latency}. The recent work\cite{yu2022metaformerbase} finds that removing the token mixer is possible but leads to performance degeneration. Those explorations in efficient token mixers inspire us to think that \textit{can we keep the vision backbone effective but removing the token mixer?} The resulting token mixer free vision backbone is expected to be efficient and effective for the realistic application.

In this work, we first review the current model architectures and learning paradigms. Most of the previous works concentrate on the improvement of the architecture while adopting the conventional supervised learning to optimize the model from scratch. Differently, we propose to adopt the simplified model architecture, and explore the learning paradigm design to fully exploit the potential of the simple model. 
We aim to simultaneously maintain the efficiency and efficacy of token mixer free vision backbone (namely IdentityFormer, in Fig.~\ref{fig:latency}-(b)). To this end, we investigate the simple and yet effective learning strategy, \textit{knowledge distillation (KD)} ~\cite{hinton2015distilling} thoroughly in the following sections. 

Our main idea is distilling the knowledge from powerful teacher model (with token mixer) to the student model (token mixer free). We instantiate the re-parameterizing idea to enlarge the modeling capacity of student network but retain its efficiency, as shown in Fig.~\ref{fig:riformer}. Specifically, the simple affine transformation is introduced into student model, to replace the token mixer for  training. The parameters of affine transformation can be merged into LayerNorm~\cite{ba2016layer} during inference, which makes the student token mixer free finally.

We empirically summarize the our learning strategy as the following guidelines, hope to shed light on how to learn the extremely simple model. Concretely,
\textbf{1}) soft distillation without using ground-truth labels is more effective; \textbf{2}) using affine transformation without distillation is difficult to tailor the performance degeneration; \textbf{3}) the proposed block-wise knowledge distillation, called \textit{module imitation}, helps leveraging the modeling capacity of affine operator; \textbf{4}) teacher with large receptive field is beneficial to improve receptive field limited student; \textbf{5}) loading the pre-trained weight of teacher model (except the token mixer) into student improve the convergence and performance.

Based on the above guidelines, we finally obtain a token mixer free vision model with competitive performance enjoying the high efficiency, dubbed as \textit{\underline{R}ep\underline{I}dentity\underline{Former} (RIFormer)}. RIFormer shares nearly the same macro and micro design as MetaFormer~\cite{yu2022metaformer}, but safely removing all token mixers. The quantitative results show that our networks outperform many prevailing backbones with faster inference speed on ImageNet-1K~\cite{deng2009imagenet}. And the ablative analyses on the feature distribution and \textit{Effective Receptive Fields (ERFs)} also demonstrate that the inductive bias brought by an explicit token mixer, can be implicitly incorporated into the simple network structure with appropriate optimization strategies. 
In summary, the main contributions of our work are as the following:

\begin{itemize}[leftmargin=*,topsep=0pt,itemsep=0pt,noitemsep]
\item We propose to explore the vision backbone by developing advanced learning paradigm for \textit{simple model architecture}, to satisfy the demand of realistic application.
\item We instantiate the re-parameterizing idea to build a token mixer free vision model, RIFormer, which owns the improved modeling capacity for the inductive bias while enjoying the efficiency during inference.
\item Our proposed practical guidelines of distillation strategy has been demonstrated effective in keeping the vision backbone competitive but removing the token mixer.
\end{itemize}
\section{Related Work}
\label{sec:related}

\subsection{Vision Transformer Acceleration}
Vision transformer is a inference slow, energy intensive backbone due to its quadratic computational cost of the self-attention, and therefore unfriendly to deploy on resource-limited edge devices, calling for compression techniques. Various technology route are designed for vision transformer slimming, such as distilling an efficient transformer with fewer depths and embedding dimensions~\cite{deit, zhang2022minivit, wu2022tinyvit, hao2022learning, wang2022attention}, pruning or merging unimportant tokens~\cite{rao2021dynamicvit, bolya2022token, meng2022adavit, kong2022spvit}, applying energy efficient operations~\cite{shu2021adder, liu2022ecoformer}, or designing efficient attention alternatives~\cite{liu2021swin, rao2021global, cai2022efficientvit}, \etc. Different from these lines, our work propose a novel angle of totally removing the complicated token mixer in a backbone while keep satisfactory performance. 

\subsection{Structual Re-parameterization}
Structual re-parameterization~\cite{ding2021repvgg, ding2022scaling, zagoruyko2017diracnets} is a training technique which decouples the training-time and inference-time architectures. For example, RepVGG~\cite{ding2021repvgg} is a plain VGG-style architecture with attractive performance and speed during inference, and a powerful architecture with manually added $1\times1$ branch and identity mapping branch during training. 
% The optimized weights of the two extra branches are equivalently merged into the $3\times3$ kernel to correlate model of training and deploy. 
Similarly, such technique can be further extended to super large kernel ConvNets~\cite{ding2022scaling}, MLP-like models~\cite{ding2022repmlpnet}, network pruning~\cite{ding2021resrep} and special optimizer design~\cite{ding2022re}. In this paper, we follow the technique to introduce parameters and equivalently absorb them into LN layer after training. The extra weights after proper optimization can help the model learn useful representations.

\section{Preliminary and Motivation}
\label{sec:preliminaries}

In this section, we first briefly recap the concept of token mixer. Then, we revisit their inevitable side effects on inference speed through an empirical latency analysis, and thus introduce the motivation of our paper. 

\begin{figure}[t]
	\centering
	\includegraphics[width=0.9\linewidth]{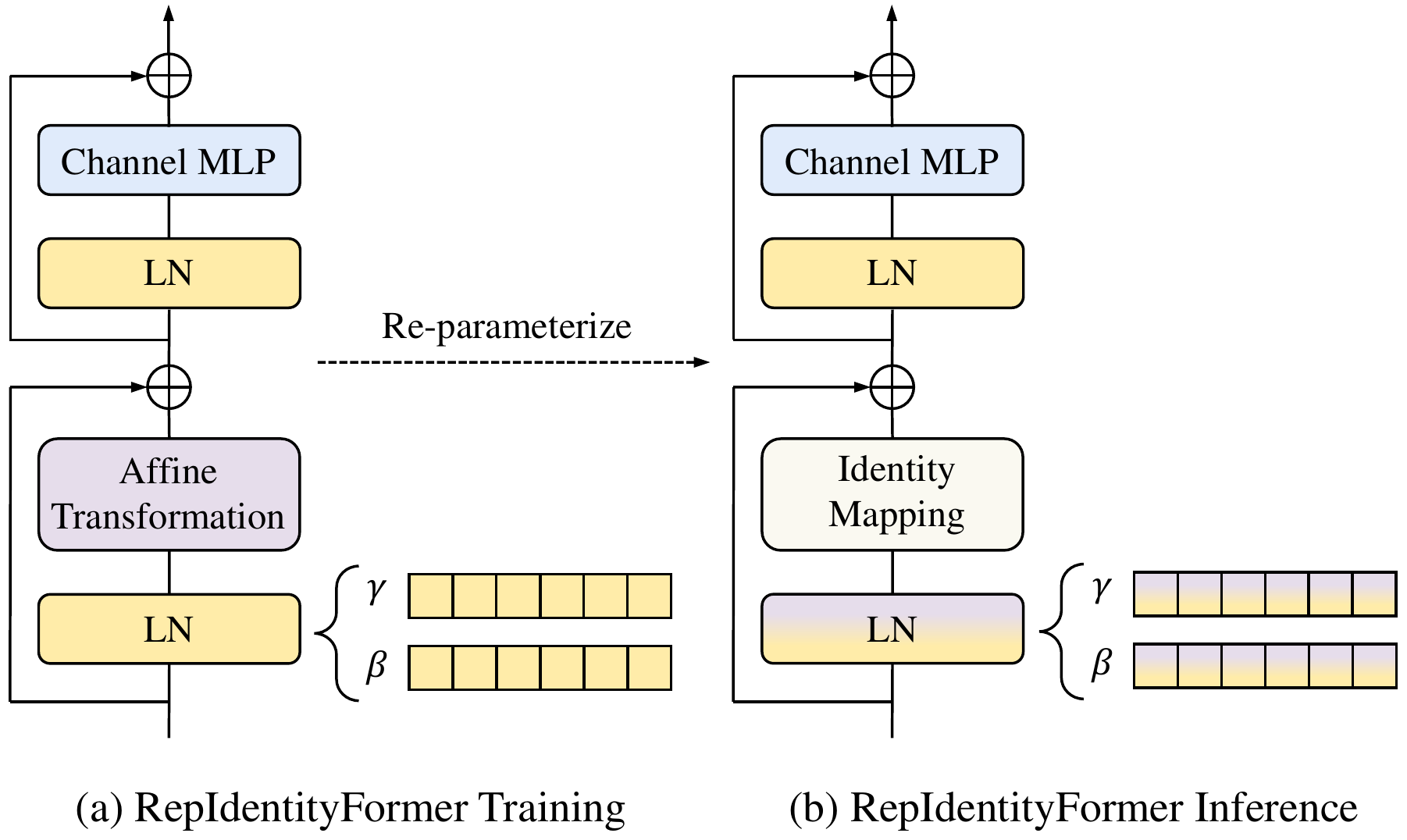}
    \vspace{-0.1in}
	\caption{Structural re-parameterization of a RIFormer block.}
	\label{fig:riformer}
    \vspace{-0.25in}
\end{figure}

\subsection{Preliminary: The Concept of Token Mixer}
The concept token mixer is a structure that perform token mixing functions in a given vision backbone. It allows information aggregation from different spatial positions~\cite{yu2022metaformer}. For instance, self-attention module serves as the token mixer in ViT~\cite{dosovitskiy2020image} by performing the attention function in parallel between components in queries, keys and values matrices, which are linearly projected from the input feature. Moreover, ResMLP~\cite{touvron2022resmlp} applies a cross-patch linear sublayer by treating Spatial MLP as token mixer. The computational and memory costs of the aforementioned token mixers are quadratic to the image scale.

\subsection{Motivation}
\label{sec:preliminaries.3}

In this section, we take our eyes on the side effects of token mixers through a quantitative latency analysis on the ViT~\cite{dosovitskiy2020image} model. We start with a modified 12-layer ViT-Base model containing only input embedding, without any operation in each of its basic building blocks. Then we gradually add the operation component (\eg, LN, Attention, Channel MLP, \etc) to each basic block, and the model finally comes to ViT-Base without the global average pooling layer and the classifier head. For each model, we take a batch size of 2048 at $224^2$ resolution with one A100 GPU and calculate the average time over 30 runs to inference that batch. The whole process is repeated for three times and we take the medium number as the statistical latency. As shown in Fig.~\ref{fig:latency}, after stacking the regular number of 12 layers, token mixer can bring an additional latency of 1433.6ms, occupying about 46.3\% of the backbone. 

According to the above analysis, token mixer brings significant side effects on the latency to the model, which makes it limited for realistic application. The observation naturally raises a question: \textit{can we keep the backbone effective but removing token mixer?}
Specifically, a recent work ~\cite{yu2022metaformerbase} introduces the MetaFormer model without any token mixer in its basic building block and finds that it raises a non-negligible performance degeneration. Based on those findings, we propose to exploit the \textit{full potential} of the extremely simple model by incorporating the inductive bias with the advanced optimization strategies, such as \textit{knowledge distillation}~\cite{hinton2015distilling, deit, zhang2022minivit}, structural re-parameterization~\cite{ding2021repvgg, ding2022scaling}, \etc. And we present all the exploration details in the remaining of this work.

\section{Exploring RIFormer: A Roadmap}
\label{sec:method}
\begin{figure*}[t]
	\centering
	\includegraphics[width=1.0\linewidth]{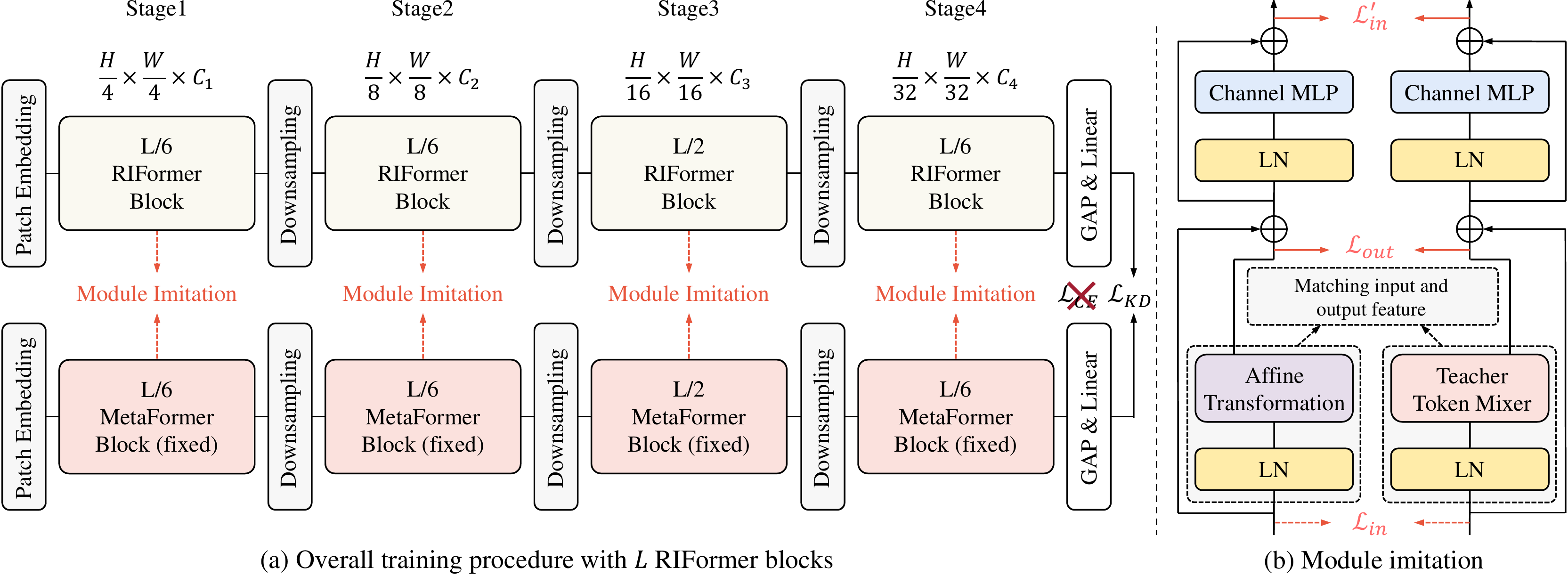}
    \vspace{-0.2in}
	\caption{\textbf{(a) Overall training procedure of RIFormer.} Following the macro and micro design of~\cite{yu2022metaformer}, RIFormer removes token mixer in each block. \textbf{(b) Module imitation technique} aims to mimic the behavior of token mixer via a simple affine transformation.}
	\label{fig:architecture}
    \vspace{-0.2in}
\end{figure*}
In this section, we present a trajectory going from a fully supervised approaches for RIFormer to more advanced training paradigms. During the journey we investigate and develop different optimization schemes for transformer-like models, while maintaining the inference-time model as the same. The baseline RIFormer we use has exactly the same macro architecture and model size as recently-developed MetaFormer~\cite{yu2022metaformer}, the difference only lies in the fact that no token mixer is used in its basic building blocks during inference. We control the computational complexity of RIFormer-S12 models comparable to PoolFormer-S12~\cite{yu2022metaformer}, with about 12M parameters and 1.8G MAC. All RIFormer-S12 models in this section are trained and evaluated on ImageNet-1K for 120 epochs. The details of hyper-parameters are shown in Sec.1 of the appendix. The roadmap of our exploration is as follows. 

\begin{table}
\vspace{0.04in}
    \begin{center}
        \small
        \begin{tabular}{lcccc}
            \hline
            Token Mixer & Training recipe   & \makecell{ImageNet top-1 acc (\%)}	\\
            \hline
            \gr 
            Pooling		&	CE Loss		&	\textbf{75.01}		\\
            \hline
            Identity    &	CE Loss		&	72.31		\\
            \hline
        \end{tabular}
    \end{center}
    \vspace{-0.2in}
    \caption{Results of different token mixers on MetaFormer using cross-entropy loss.}
    \vspace{-0.1in}
    \label{table-4-1}
\end{table}

\begin{table}
    \begin{center}
        \small
        \begin{tabular}{lcccc}
            \hline
            TM & Label  & Teacher  & ImageNet top-1 acc (\%)	\\
            \hline
            Identity    & \ding{51}\xspace & \ding{55}\xspace & 72.31  \\
            \hline
            Identity    & \ding{51}\xspace & hard & 73.51  \\

            Identity    & \ding{55}\xspace & hard & 72.86  \\
            \hline
            Identity    & \ding{51}\xspace & soft & 73.64  \\

            \gr
            Identity    & \ding{55}\xspace & soft & \textbf{74.05}  \\
            \hline
        \end{tabular}
    \end{center}
    \vspace{-0.2in}
    \caption{Results of different teacher type in normal/label-free RIFormer-S12 with identity mapping as token mixer.}
    \label{table-4-2}
    \vspace{-0.1in}
\end{table}

\subsection{Vision Backbone Without Token Mixer} 
\label{sec:method.1}
Our exploration is directed to remove token mixer in each basic block of a inference-time model vision backbone to obtain a higher inference speed while striving to keep the performance. Thus, we start with a RIFormer-S12 model with a fully supervised training scheme using CE loss, mainly follows~\cite{yu2022metaformer}. As a performance reference, we compare the results with PoolFormer-S12, since it use only basic pooling operation as token mixer and the performance gap can thus, be attributed to the absence of basic token mixing function. As shown in Tab.~\ref{table-4-1}, RIFormer-S12 with a trivial supervised training can lead to an unacceptable performance drop ($2.7\%$ top-1 accuracy) compared to PoolFormer-S12. The results show that without token mixer in each building block, it is limited for regular supervised learning in helping the model learn useful information from images, calling for advanced training procedure. 

\begin{table}
    \begin{center}
        \small
        \begin{tabular}{lcccc}
            \hline
            TM & Label  & KD type  & \makecell{ImageNet top-1 acc (\%)}	\\
            \hline
            Affine    & \ding{51}\xspace & \ding{55}\xspace & 72.25  \\
            \hline
            Affine    & \ding{51}\xspace & hard & 73.44  \\

            Affine    & \ding{55}\xspace & hard & 72.77  \\
            \hline
            Affine    & \ding{51}\xspace & soft & 72.10  \\

            \gr
            Affine    & \ding{55}\xspace & soft & \textbf{74.07}  \\
            \hline
            
        \end{tabular}
    \end{center}
    \vspace{-0.2in}
    \caption{Results of different distillation type in normal/label-free RIFormer-S12 with affine transformation as token mixer.}
    \label{table-4-3}
    \vspace{-0.2in}
\end{table}

We then investigate and modify a series of training paradigms to improve the inferior baseline performance, which can be summarized as 1) knowledge distillation, 2) teacher type influence, 3) structural re-parameterization, 4) the proposed module imitation technique, 5) load partial parameters from teacher. Since we aim at exploring the influence of different advanced training recipes instead of network architecture, inference-time model architecture is always kept the same at intermediate steps. Next, we share 5 useful guidelines for training RIFormer. 

\subsection{Distillation Paradigm Design} 
\label{sec:method.2}
We now study the knowledge distillation~\cite{hinton2015distilling, deit} of a RIFormer student by a general vision backbone teacher with token mixer, and summarize how to effectively utilize the "soft" labels coming from the strong teacher network.
\vspace{-0.2in}
\paragraph{Guideline 1: soft distillation without using ground-truth labels can be effective for student without token mixer.} Basically, most of the existing KD methods are designed for models with token mixer. For example, it is common practice to help a student \textit{convnet} by learning from both ground-truth labels and the soft labels predicted by a teacher convnet. Moreover, some observations from DeiT~\cite{deit} show that using the hard labels instead of soft labels as a supervised target, can improves \textit{transformer} significantly. In contrast, the token mixer free backbone do not have explicit patch aggregating modules in its basic block. The distillation of it is should be thus, different from that of conventional backbones. Specifically, although RIFormer shares the same macro structure as transformer, it still cannot be treated as a student \textit{transformer} because we have deliberately removed the token mixer from each building block. However, we also do not prefer viewing it as a pure \textit{convnet} since RIFormer bears a resemblance to transformer in terms of macro/micro-level architecture design. Therefore, we are motivated to explore a suitable KD method for RIFormer with promising performance. 

Typically, the cross-entropy objective is to assist a student network reproduce the hard accurate label, and we argue that the process may be unsuitable for RIFormer. First, the ground-truth hard label can be transformed to a soft distribution by label-smoothing regularization~\cite{szegedy2016rethinking}, with weights $1-\varepsilon$ for the true label and $\varepsilon$ shared each classes. The unlearned uniform distribution across the negative classes is less informative, and may interfere with the learned soft distribution given by teacher. Second, 1$\times$1 convolutions actually dominate basic building block in RIFormer, "mixing" only the per-location features but not spatial information. Such a simplified design may require richer information in the supervised labels. To demonstrate this, Tab.~\ref{table-4-2} compare the performance of four different settings. The default teacher is a GFNet-H-B~\cite{rao2021global} (54M parameters). Hard distillation with true labels improve the accuracy from $72.31\%$ to $73.51\%$. It shows that a teacher with token mixer has a positive effect on a student without token mixer.  In fact, the combination of using a soft distillation without true labels performs the best, improving the network performance to $74.05\%$.

\noindent \textbf{Remark 1.} Supervised learning with true label does not seem to be the most suitable way for a crude model without token mixer. A teacher with token mixer can help to guide the training, but still fails to fully recover the performance gap from removing token mixer, calling for other strategies. 

\subsection{Re-parameterization for Identity Mapping} 
\label{sec:method.3}

\paragraph{Guideline 2: using affine transformation without tailored distillation, is hard to recover the performance degredation.} In this part, we adopt the idea of \textit{Structural Reparameterization}~\cite{ding2021repvgg, ding2022scaling, ding2022repmlpnet} methodology, which usually takes a powerful model for training and equivalently converts to a simple model during inference. Specifically, the inference-time token mixer module in RIFormer can be viewed as an identity mapping following a LN layer. Thus, the training-time module should satisfy at least two basic requirements: 1) \textit{per-location} operator for allowing equivalent transformation; 2) \textit{parametric} operator for allowing extra representation ability. Accordingly, we apply an affine transformation operator to replace the identity mapping during training, which only performs channel-wise scaling and shifts, as shown in Fig.~\ref{fig:riformer}. The affine operator and its preceding LN layer can be converted into a LN with modified weights, thus it can be equivalently converted into an identity mapping during inference. Denote the input feature as $\mathrm{M}\in\mathbb{R}^{N\times C\times H\times W}$, the affine operator can be expressed as: 
\begin{table}
    \begin{center}
        \small
        \begin{tabular}{lccccc}
            \hline
            TM & Feat  & Rel  & Layer  & \makecell{ImageNet top-1 acc (\%)}	\\
            \hline
            Affine   & 0 & 0 & - & 74.07  \\
            \hline
            Affine   & 40 & 0 & 6 & 74.49  \\
            
            Affine   & 60 & 0 & 6 & 74.77  \\
            
            Affine   & 80 & 0 & 6 & 74.81  \\
            
            % Affine$\times3$   & 80 & 0 & 6 & 74.77  \\

            % Affine$\times9$   & 80 & 0 & 6 & 74.74  \\
            \hline
            Affine   & 80 & 10 & 6 & 75.08  \\
            
            Affine   & 80 & 20 & 6 & 74.82  \\
            
            Affine   & 80 & 40 & 6 & 75.00  \\

            \gr
            Affine   & 80 & 20 & 4 & \textbf{75.13}  \\
            \hline
        \end{tabular}
    \end{center}
    \vspace{-0.2in}
    \caption{Results of different module imitation setting.}
    \label{table-4-4}
    \vspace{-0.2in}
\end{table}
\vspace{-0.2in}
\begin{equation}
\begin{aligned}
\mathrm{Affine}(\mathrm{M},\boldsymbol{\mathbf{s}},\boldsymbol{\mathbf{t}})_{:,i,:,:} & = \boldsymbol{\mathbf{s}}_{i} \mathrm{M}_{:,i,:,:} + \boldsymbol{\mathbf{t}}_{i} - \mathrm{M}_{:,i,:,:}, 
\end{aligned}
\label{equ:method_1}
\end{equation}
where $\mathbf{s}\in\mathbb{R}^{C}$ and $\mathbf{t}\in\mathbb{R}^{C}$ are learnable weight vectors. We follow~\cite{yu2022metaformer} to add a subtraction of the input during implementation due to the residual connection, and thus does not merge the first and third terms in Eq.~\ref{equ:method_1}. Then, we describe how to merge the affine transformation into its preceding LN layer, so the training-time model can be equivalently converted to model for deploy but no longer has token mixer in its blocks. We use $\vect{\mu},\vect{\sigma},\vect{\gamma},\vect{\beta}$ as the mean, standard deviation and learned scaling factor and bias of the preceding LN layer. Denote $\mathrm{T}^{(a)}\in\mathbb{R}^{N\times C\times H\times W}$, ${\mathrm{T}^{\prime(a)}}\in\mathbb{R}^{N\times C\times H\times W}$ respectively as the input and output of an affine residual sub-block in Fig.~\ref{fig:riformer}-(a). During the training time, we have:
% \vspace{-0.15in}
\begin{equation}
\begin{aligned}
\mathrm{T}^{\prime(a)} &= \mathrm{Affine}(\mathrm{LN}(\mathrm{T}^{(a)},\vect{\mu},\vect{\sigma},\vect{\gamma},\vect{\beta}),\boldsymbol{\mathbf{s}},\boldsymbol{\mathbf{t}}) -\mathrm{T}^{(a)}
\end{aligned}
\label{equ:method_2}
\end{equation}
where $\mathrm{LN}$ is the LN function, which is implemented by GroupNorm API in PyTorch (setting the group number as 1) following~\cite{yu2022metaformer}. During inference time, there only exists an identity mapping followed by a LN layer in the residual sub-block. Thus, we have:
\begin{equation}
\begin{aligned}
\mathrm{T}^{\prime(a)} &= \text{LN}(\mathrm{T}^{(a)},\vect{\mu},\vect{\sigma},\vect{\gamma}^\prime,\vect{\beta}^\prime),
\end{aligned}
\label{equ:method_3}
\end{equation}
where $\vect{\gamma}^\prime$ and $\vect{\beta}^\prime$ are the weight and bias parameters of the merged LN layer. Based on the equivalence of Eq.~\ref{equ:method_2} and Eq.~\ref{equ:method_3}, for $\forall 1\leq i \leq C$, we have:
\begin{equation}
\begin{aligned}
\vect{\gamma}^\prime_i = \vect{\gamma}_i(\boldsymbol{\mathbf{s}}_i-1),~~~ 
\vect{\beta}^\prime_i = \vect{\beta}_i(\boldsymbol{\mathbf{s}}_i-1)  +\boldsymbol{\mathbf{t}}_i, 
\end{aligned}
\label{equ:method_4}
\end{equation}
The proof and PyTorch-like code of the affine transformation and re-parameterization process is shown in Sec.2 and Sec.3 of the appendix, respectively. Since the LN layer does not have a pre-computed mean and standard deviation during inference time, their specific values are input adaptive, which do not affect the equivalence of transform.

\noindent \textbf{Remark 2.} Compare Tab.~\ref{table-4-3} with Tab.~\ref{table-4-2}, directly applying structural re-parameterization method shows no advantages. We attribute this phenomenon to the fact that the affine transformation in the LN layer is a \textit{linear transformation} that can be directly merged with the extra affine operator we introduced (if do not add any nonlinear function in between). Therefore, if both are supervised only by the output of the model, the potential of the additional parameters may not be fully exploited. Meanwhile, the isomorphic design of teacher and student inspires us to explore suitable methods for knowledge transfer of \textit{modules} at each layer.

\begin{table}
    \begin{center}
        \small
        \begin{tabular}{lccc}
            \hline
            Teacher (T)  & T.acc (\%) & MI  & \makecell{ImageNet \\ top-1 acc (\%)}	\\
            \hline
            PoolFormer-M48~\cite{yu2022metaformer} & 82.5 & \ding{55}\xspace & 73.63  \\
            
            Swin-B$^*$~\cite{liu2021swin} & 85.2 & \ding{55}\xspace & 73.12  \\
            
            Pyramid ViG-B~\cite{han2022vision} & 83.7 & \ding{55}\xspace & 73.25  \\
            
            \gr
            GFNet-H-B~\cite{rao2021global} & 82.9 & \ding{55}\xspace & 74.07  \\

            \hline
            PoolFormer-M48~\cite{yu2022metaformer} & 82.5 & \ding{51}\xspace & 74.83  \\
            
            Swin-B$^*$~\cite{liu2021swin} & 85.2 & \ding{51}\xspace & 74.52  \\
            
            Pyramid ViG-B~\cite{han2022vision} & 83.7 & \ding{51}\xspace & 74.25  \\
            
            \gr
            GFNet-H-B~\cite{rao2021global} & 82.9 & \ding{51}\xspace & \textbf{75.13}  \\
            \hline
            
        \end{tabular}
    \end{center}
    \vspace{-0.2in}
    \caption{Results of different teachers on RIFormer-S12 w/ or w/o module imitation (MI). $^*$ indicates ImageNet-22K pre-training.}
    \label{table-4-5}
    \vspace{-0.2in}
\end{table}

\subsection{Module Imitation} 
\label{sec:method.4}

\paragraph{Guideline 3: the proposed block-wise knowledge distillation, called module imitation, helps leveraging the modeling capacity of affine operator.} The previous KD methods we tried only focus on the output of between teacher and student networks. We propose \textit{module imitation (MI)} method, which present to utilize the useful information in the teacher's token mixer. Specifically, a pretrained PoolFormer-S12~\cite{yu2022metaformer} is utilized as a teacher network.
As shown in Fig.~\ref{fig:architecture}, we expect the simple affine operator (with its preceding LN layer) to approximate the behavior of that of a basic token mixer during training. Denote $f(\cdot), \mathrm{T}^{(a),m}\in\mathbb{R}^{N\times C\times H\times W} ,m\in \mathcal{M}$ as the affine operator and the input of the $m$-th layer of RIFormer in which $\mathcal{M}$ is the intermediate layers set we used, and $g(\cdot), \mathrm{T}^{(t),m}\in\mathbb{R}^{N\times C\times H\times W}, m\in \mathcal{M}$ are that of the teacher network, respectively. We abbreviate $\mathrm{LN}(\cdot,\vect{\mu},\vect{\sigma},\vect{\gamma},\vect{\beta})$ as $\mathrm{LN}(\cdot)$ for simplicity. The mean squared error (MSE) of the inputs between the LN layer of affine operator and token mixer can be calculated as:
\begin{equation}
\begin{aligned}
\mathcal{L}_{in}=\alpha_1\Vert \mathrm{LN}(\mathrm{T}^{(a),m}) - \mathrm{LN}(\mathrm{T}^{(t),m}) \Vert_F^2,
\end{aligned}
\label{equ:method_5}
\end{equation}
where $\alpha_1=1/NCHW$. Note that the input feature of the current layer is the output feature of the previous one. Therefore, we propose to match the output features of this block (\ie, the input features of the next subsequent block) in practice, which can be seen as a hidden state distillation in transformers~\cite{jiao2019tinybert, hao2022learning, zhang2022minivit, wang2020minilm, wang2020minilmv2}.
\begin{equation}
\begin{aligned}
\mathcal{L}_{in}^\prime=\alpha_1\Vert \mathrm{T}^{(a),m+1} - \mathrm{T}^{(t),m+1} \Vert_F^2,
\end{aligned}
\label{equ:method_6}
\end{equation}
The hidden-state distillation based on relation matrices~\cite{zhang2022minivit, hao2022learning} is then applied on the output feature:
\begin{equation}
\begin{aligned}
\mathcal{L}_{rel}=\alpha_2\Vert \mathcal{R}(\mathrm{T}^{(a),m+1}) - \mathcal{R}(\mathrm{T}^{(t),m+1}) \Vert_F^2, 
\end{aligned}
\label{equ:method_7}
\end{equation}
where $\alpha_2=1/NH^2W^2$, $\mathcal{R}(T)=\tilde{T}\tilde{T}^\mathsf{T}$, $\tilde{T}$ denotes normalize $T$ at the last dimension. Considering the MSE of the outputs between affine operator and token mixer:
\begin{equation}
\begin{aligned}
\mathcal{L}_{out}=\alpha_1\Vert f(\mathrm{LN}(\mathrm{T}^{(a),m})) - g(\mathrm{LN}(\mathrm{T}^{(t),m})) \Vert_F^2, 
\end{aligned}
\label{equ:method_8}
\end{equation}
Combining Eq.~\ref{equ:method_6}, Eq.~\ref{equ:method_7} and Eq.~\ref{equ:method_8},  the final loss function with module imitation is defined as:
\begin{equation}
\begin{aligned}
\mathcal{L}=\mathcal{L}_{soft} + \lambda_1\mathcal{L}_{in}^\prime + \lambda_2\mathcal{L}_{out} + \lambda_3\mathcal{L}_{rel},
\end{aligned}
\label{equ:method_9}
\end{equation}
where $\mathcal{L}_{soft}$ is the soft logit distillation target in Sec.~\ref{sec:method.2}, $\lambda_1$, $\lambda_2$, $\lambda_3$ is the hyper-parameter for seeking the balance between loss functions. In Tab.~\ref{table-4-4}, Feat  and Rel are number of epochs of using $(\mathcal{L}_{in}^\prime, \mathcal{L}_{out})$ and $\mathcal{L}_{rel}$, Layer represents the number of intermediate layers we used. The results show positive effect of module imitation on the student RIFormer in different circumstances. With a 4 layer setting and the usage of affine operator, we get the best result of $75.13\%$, already surpassing the PoolFormer-S12's result of $75.01\%$ in Tab.~\ref{table-4-1}. From now on, we will use this setting. 

\noindent \textbf{Remark 3.} We deem a reason for that phenomenon might be that module imitation helps the affine operator implicitly benefit from the supervision of the teacher's token mixer, while not losing the convenience of explicitly merging the preceding LN layer. Besides, we find module imitation can effectively shift the feature distribution closer to the teacher network and show larger \textit{Effective Receptive Fields (ERFs)}. Please refer to Sec.~\ref{sec:experiments.3} for details. 
\vspace{-0.1in}
\begin{table*}[t]
    \centering
    \small
        \begin{tabular}{c|lccccccc}
        \toprule
    Token Mixer    & Outcome Model    & Image Size & Params (M)  & MACs (G) & Throughput (images/s) & Top-1 (\%) \\
        \whline
    \multirow{4}{*}{Convolution}   
        %  & \resnetdot{} RSB-ResNet-18 \cite{he2016deep, resnet_improved} & 224 & 12 & 1.8 & 10445.10 & 70.6 \\
         & \resnetdot{} RSB-ResNet-34 \cite{he2016deep, wightman2021resnet} & 224 & 22 & 3.7 & 6653.75 & 75.5 \\
         & \resnetdot{} RSB-ResNet-50 \cite{he2016deep, wightman2021resnet} & 224 & 26 & 4.1 & 2732.85 & 79.8 \\
         & \resnetdot{} RSB-ResNet-101 \cite{he2016deep, wightman2021resnet} & 224 &  45 & 7.9 & 1856.48 & 81.3 \\
         & \resnetdot{} RSB-ResNet-152 \cite{he2016deep, wightman2021resnet} & 224 & 60 & 11.6 & 1308.26 & 81.8 \\
    \hline
         
    \multirow{5}{*}{Attention} 
            %  & \vitdot{} ViT-B/16$^*$ \cite{dosovitskiy2020image} & 224 &  86 & 17.6 & - & 79.7 \\
            %  & \vitdot{} ViT-L/16$^*$ \cite{dosovitskiy2020image} & 224 & 307 & 63.6 & - & 76.1 \\
             & \deitdot{} DeiT-S \cite{deit} & 224 & 22 & 4.6 & 3092.02 & 79.8 \\
             & \deitdot{} DeiT-B \cite{deit} & 224 &  86 & 17.5 & 1348.76 & 81.8 \\
            %  & \pvtdot{} PVT-Tiny \cite{wang2021pyramid} & 224 & 13 & 1.9 & 2726.65 & 75.1 \\
             & \pvtdot{} PVT-Small \cite{wang2021pyramid} & 224 & 25 & 3.8 & 1622.53 & 79.8 \\
             & \pvtdot{} PVT-Medium \cite{wang2021pyramid} & 224 & 44 &  6.7 & 1190.48 & 81.2 \\
             & \pvtdot{} PVT-Large \cite{wang2021pyramid} & 224 & 61 &  9.8 & 865.33 & 81.7 \\
    \hline
    \multirow{5}{*}{Spatial MLP} 
             & \mlpmixerdot{} MLP-Mixer-B/16 \cite{tolstikhin2021mlp} & 224 & 59 & 12.7 & 1855.45 & 76.4 \\
            %  & \resmlp{} ResMLP-S12 \cite{touvron2022resmlp} & 224 & 15 & 3.0 & 6299.31 & 76.6 \\
             & \resmlp{} ResMLP-S24 \cite{touvron2022resmlp} & 224 & 30 & 6.0 & 3228.75 & 79.4 \\
             & \resmlp{} ResMLP-B24 \cite{touvron2022resmlp} & 224 & 116 & 23.0 & 298.94 & 81.0 \\
            %  & \swinmixer{} Swin-Mixer-T/D24 \cite{liu2021swin} & 256 & 20 & 4.0 & 1673.21 & 79.4 \\
             & \swinmixer{} Swin-Mixer-T/D6 \cite{liu2021swin} & 256 & 23 & 4.0 & 1625.59 & 79.7 \\
             & \swinmixer{} Swin-Mixer-B/D24 \cite{liu2021swin} & 224 & 61 & 10.4 & 1131.60 & 81.3 \\
            %  & \gmlp{} gMLP-S \cite{gmlp} & 224 & 20 & 4.5 & - & 79.6 \\
            %  & \gmlp{} gMLP-B \cite{gmlp} & 224 & 73 & 15.8 & - & 81.6 \\
    \hline
    \multirow{4}{*}{2D FFT}  
             & {\color{green}$\blacksquare$} GFNet-H-Ti~\cite{rao2021global} & 224 & 15 & 2.1 & 1979.56 & 80.1 \\ 
             & {\color{green}$\blacksquare$} GFNet-H-S~\cite{rao2021global} & 224 & 32 & 4.6 & 1434.19 & 81.5 \\
             & {\color{green}$\blacksquare$} GFNet-B~\cite{rao2021global} & 224 & 43 & 7.9 & 1771.07 & 80.7 \\ 
             & {\color{green}$\blacksquare$} GFNet-H-B~\cite{rao2021global} & 224 & 54 & 8.6 & 939.20 & 82.9 \\ 
    \hline
    \multirow{5}{*}{Pooling}  
             & \poolformer{} PoolFormer-S12~\cite{yu2022metaformer} & 224 & 12 & 1.8 & 4160.18 & 77.2 \\ % 77.3
             & \poolformer{} PoolFormer-S24~\cite{yu2022metaformer} & 224 & 21 & 3.4 & 2140.20 & 80.3 \\
             & \poolformer{} PoolFormer-S36~\cite{yu2022metaformer} & 224 & 31 & 5.0 & 1440.37 & 81.4 \\ % 81.5
             & \poolformer{} PoolFormer-M36~\cite{yu2022metaformer} & 224 & 56 & 8.8 & 1009.45 & 82.1 \\
             & \poolformer{} PoolFormer-M48~\cite{yu2022metaformer} & 224 & 73 & 11.6 & 761.93 & 82.5 \\
    \hline
    \multirow{10}{*}{None}  
             & {\color{orange}$\bigstar$} RIFormer-S12$^\diamond$ & 224 & 12 & 1.8 & 4899.80 (\better{17.8\%}) & 76.9 \\ 
             & {\color{orange}$\bigstar$} RIFormer-S24$^\diamond$ & 224 & 21 & 3.4 & 2530.48 (\better{18.2\%}) & 80.3 \\
             & {\color{orange}$\bigstar$} RIFormer-S36$^\diamond$ & 224 & 31 & 5.0 & 1699.94 (\better{18.0\%}) & 81.3 \\ 
             & {\color{orange}$\bigstar$} RIFormer-M36$^\diamond$ & 224 & 56 & 8.8 & 1185.33 (\better{17.4\%}) & 82.6 \\
             & {\color{orange}$\bigstar$} RIFormer-M48$^\diamond$ & 224 & 73 & 11.6 & 897.05 (\better{17.7\%}) & 82.8 \\
             & {\color{orange}$\bigstar$} RIFormer-S12$^\ddagger$ & 384 & 12 & 5.4 & 1586.51 & 78.3 \\ 
             & {\color{orange}$\bigstar$} RIFormer-S24$^\ddagger$ & 384 & 21 & 10.0 & 819.40 & 81.4 \\
             & {\color{orange}$\bigstar$} RIFormer-S36$^\ddagger$ & 384 & 31 & 14.7 & 552.07 & 82.2 \\ 
             & {\color{orange}$\bigstar$} RIFormer-M36$^\ddagger$ & 384 & 56 & 25.9 & 403.15 & 83.4 \\
             & {\color{orange}$\bigstar$} RIFormer-M48$^\ddagger$ & 384 & 73 & 34.1 & 304.43 & 83.7 \\
    \bottomrule
             
    \end{tabular}
    \vspace{-3mm}
    \caption{
    \textbf{Results of models with different types of token mixers on ImageNet-1K.} 
    $^\diamond$ denotes training with prolonged 600 epochs. $^\ddagger$ denotes fine-tuning from the ImageNet pre-trained model for 30 epochs.
    \label{tab:imagenet}}
    \vspace{-6mm}
\end{table*}
\paragraph{Guideline 4: teacher with large receptive field is beneficial to improve student with limited receptive field.} Tab.~\ref{table-4-5} compares student performance with different teacher architectures. Although GFNet-H-B~\cite{rao2021global} does not show the highest ImageNet top-1 performance among teachers, it can still serves as a better choice, no matter whether module imitation is used or not. 

\noindent \textbf{Remark 4.} The fact is probably attributed to the receptive field gap between teacher and student. As explained by
~\cite{abnar2020transferring}, inductive bias can be transferred from one model to another through distillation. According to this study, model with large receptive field (\eg, GFNet with learnable \textit{global filters} in the frequency domain) can be better teacher for student RIFormer with limited receptive field. 

\paragraph{Guideline 5: loading the pre-trained weight of teacher model (except the token mixer) into student improve the convergence and performance.} Our method can be categorized as a model compression technique that aims at removing the token mixer in basic blocks for acceleration. Inspried by previous methods, including knowledge distillation~\cite{sun2019patient, sanh2019distilbert}, quantization~\cite{liu2018bi, li2022q}, and model acceleration~\cite{rao2021dynamicvit} that initialize the weights of the light-weight network using (or partly using) the corresponding weights of the pre-trained heavy network, we explore a suitable initialization method. Since our goal is to remove only the token mixer, the weights of the remaining part still remain and are not paid enough attention in the previous journey. We observe that initializing the weights of RIFormer (except the affine operator) with the corresponding teacher network further boost the performance from $75.13\%$ to $75.36\%$. This brings us to the final paradigm for training RIFormer. 
\vspace{-0.2in}
\paragraph{Closing remarks.} So far, we have finished our exploration and discovered a suitable paradigm for training the RIFormer. It has the approximately the same macro design with MetaFormer~\cite{yu2022metaformer}, but does not require any token mixer. Equipped with the propsoed optimization methods, RIFormer can outperform complicated models with token mixers for ImageNet-1K classification. These encouraging findings inspire us to answer the following questions in the next section. 1) The scaling behavior of such extremely simple architecture with our training paradigm. 2) The generalizability of the paradigm on different teachers.

\section{Experiments}
\label{sec:experiments}

\subsection{Image classification} 
\label{sec:experiments.1}
\paragraph{Setup.} For ImageNet-1K~\cite{deng2009imagenet} with $1.2$M training images and $50000$ validation images, we generally apply the training scheme in~\cite{yu2022metaformer} while following the guidelines in Sec.~\ref{sec:method}. The data augmentation contains MixUp~\cite{mixup}, CutMix~\cite{cutmix}, CutOut~\cite{cutout} and RandAugment~\cite{randaugment}. 
% We use AdamW~\cite{loshchilov2017decoupled} optimizer with a batch size of $1024$ and weight decay of $0.05$, and learning rate $\mathrm{lr} = 1e^{-3} \cdot \mathrm{batch\ size} / 1024$. Stochastic Depth\cite{stochastic_depth}, Label Smoothing\cite{label_smoothing} and Layer Scale\cite{touvron2021going} are also adopted to regularize the networks. 
As a model compression work on removing token mixer, bridging the performance gap caused by removing the token mixer is definitely our first priority, instead of proposing a strong baseline. Therefore, we use a prolonged the training epochs of $600$. We also finetune the pre-trained models for 30 epochs with input resolution of $384^2$. More details are in the appendix. 
% For each model we take a batch size of 2048 at $224^2$ resolution (1024 for $384^2$ resolution) with one 80GB A100 GPU and calculate the average throughput over 30 runs to inference that batch. The process is repeated for three times and the medium are treated as the statistical throughput. 
\vspace{-0.2in} 
\paragraph{Main results.}
Tab.~\ref{tab:imagenet} shows the results of RIFormer on ImageNet classification. We pay main attention to the \textit{throughput} metrics since our primary consideration is to satisfy latency requirements for edge devices. 
As expected, favorable speed advantage is achieved since RIFormer does not contain any token mixer in its building block, compared with other type of backbones. Surprisingly, with such fast inference, RIFormers successfully remove all token mixers using our training approach without affecting the performance. For example, RIFormer-M36 can process more than $1185$ images at $224^2$ resolution per second, with the top-1 accuracy of $82.6\%$. In comparison, the recent baseline PoolFormer-M36~\cite{yu2022metaformer} with a Pooling token mixer, can process $1009$ images of the same size with a worse $82.1\%$ accuracy. We also compare with an efficient backbone, GFNet~\cite{rao2021global}. It conducts token mixing via a global filter, which consists an FFT, an element-wise multiplication, and an IFFT, with a total computational complexity $\mathcal{O}(N\log N)$. With a $939$ throughput, GFNet-H-B gets $82.9\%$ accuracy while our RIFormer-M48 can still reaches a comparable $82.8\%$ accuracy with on par throughput of $897$. Meanwhile, the body of inference-time RIFormer is dominated by only $1\times1$ conv following LN, without complex 2D FFT or attention, friendly for hardware specialization. 

Notably, without token mixer, RIFormer cannot even perform basic token mixing operation in its building blocks. However, the ImageNet experiments demonstrate that with the proposed training paradigm, RIFormer still shows promising results. We can only deem the reason behind the fact might be that optimization strategy plays a key role. RIFormer is readily a starting recipe for the exploration of optimization-driven efficient network design, and rest assured of the performance with advanced training schemes. 

\subsection{Ablation studies}
\label{sec:experiments.2}
\paragraph{Effectiveness of module imitation.} As an important way for the extra affine operator to learn suitable weights, module imitation is based on distillation. Therefore, we compare it with the hidden state feature distillation approach (with relations). Taking the paradigm in Sec.~\ref{sec:method.2} by soft distillation without CE loss, we get the results in Tab.~\ref{table-ab-1}. More details for Sec.~\ref{sec:experiments.2} can be found in Sec.4 of the appendix. With feature distillation, the accuracy is $0.46\%$ lower than that of module imitation, showing module imitation's positive effect on the optimization of the extra weights. 
\begin{figure}[t]
	\centering
	\includegraphics[width=1.0\linewidth]{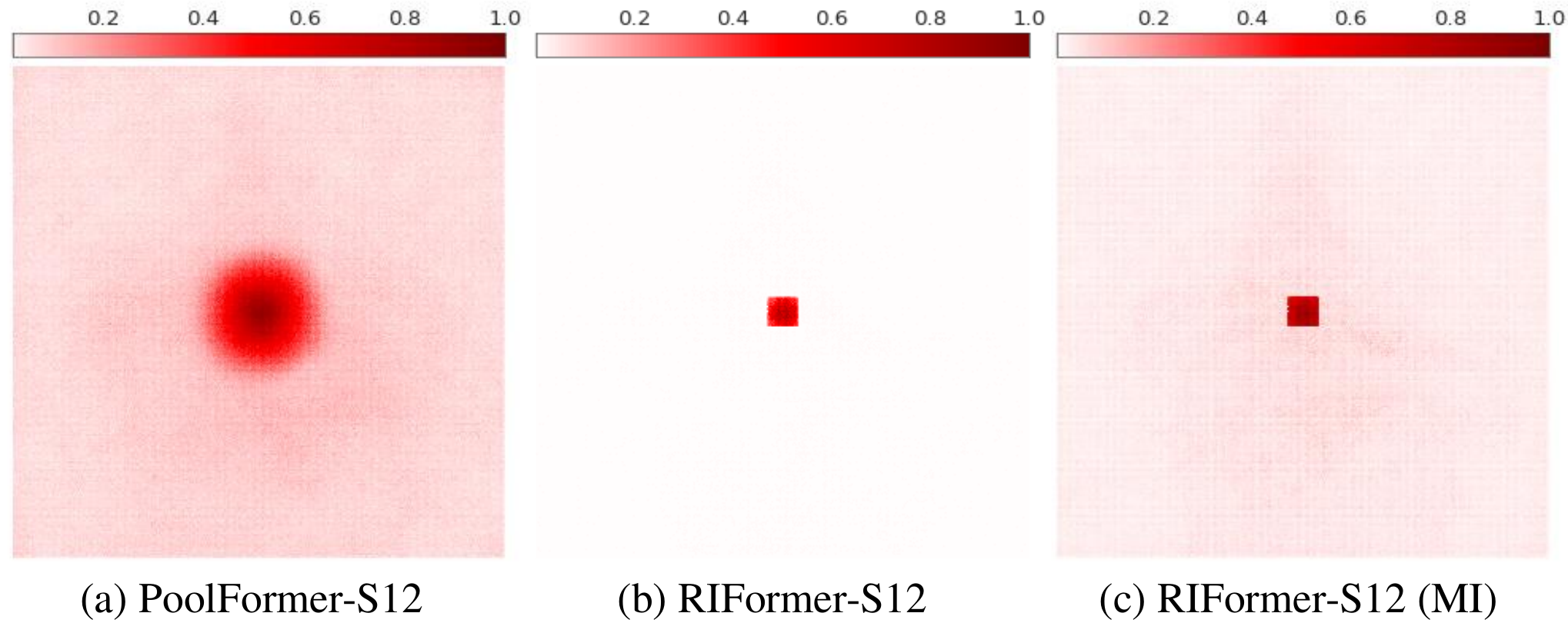}
    \vspace{-0.2in}
	\caption{The \textit{Effective Receptive Field (ERF)} of PoolFormer-S12 and RIFormer-S12 with/without using module imitation.}
	\label{fig:erf}
    \vspace{-0.1in}
\end{figure}

\begin{table}[t]
% \vspace{0.04in}
    \begin{center}
        \small
        \begin{tabular}{lcccc}
            \hline
            Token Mixer & Feature distillation scheme  &   Top-1 (\%)	\\
            \hline
            Identity    &	None 	&	74.05		\\
            \hline
            Identity    &	Feature distill		&	74.90		\\
            \gr
            Affine		&	Module imitation		&	\textbf{75.36}		\\
            \hline
        \end{tabular}
    \end{center}
    \vspace{-0.2in}
    \caption{Ablation study of the effectiveness of module imitation.}
    \label{table-ab-1}
    \vspace{-0.22in}
\end{table}

\vspace{-0.15in}
\paragraph{Comparisons of different acceleration strategy.}
Next, we discuss whether the token removing is better than other sparsification strategies. Based on the PoolFormer~\cite{yu2022metaformer} baseline, we first construct a slim PoolFormer-S9 and PoolFormer-XS12 by reducing the depth to $9$ and by maintaining about $\frac{5}{6}$ of its width, \ie, embedding dimension, to obtain comparable inference speed with our RIFormer-S12. We also follow the soft distillation paradigm in Sec.~\ref{sec:method.2}. Tab.~\ref{table-ab-2} shows the results. Directly pruning depths or width cannot render a better performance than ours without latency-hungry token mixer. 

\vspace{-0.15in} 
\paragraph{Generalization to different teachers.} 
In order to verify the proposed training paradigm a general compression technique, we adopt the architecture modifications in~\cite{yu2022metaformerbase} for student and change teacher to the other 4 MetaFormer baselines~\cite{yu2022metaformerbase}, with teacher token mixer as rand matrices, pooling, separable depthwise convolutions, and attention, respectively. Tab.~\ref{table-ab-3} shows that our method has a positive effect in different depth settings and teacher circumstances. 

\subsection{Analysis of Module Imitation.}
\label{sec:experiments.3}
\begin{figure}[t]
	\centering
	\includegraphics[width=1.0\linewidth]{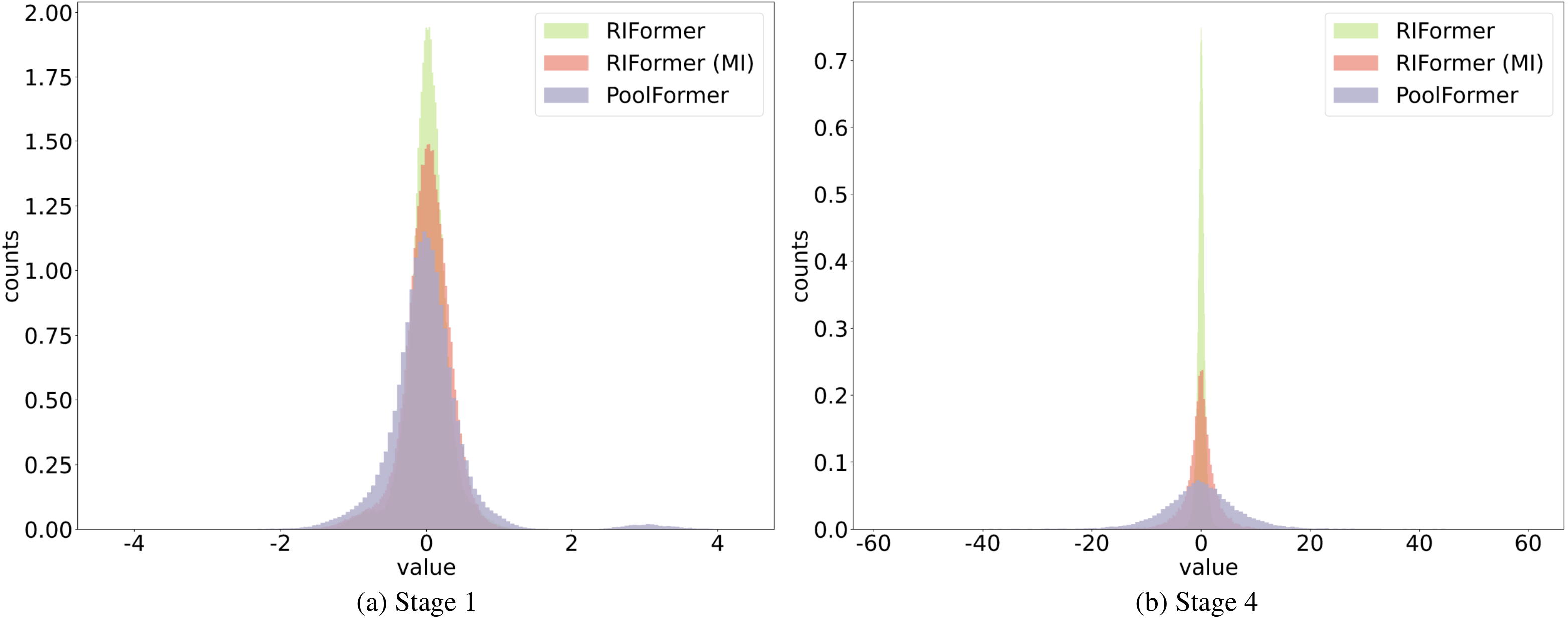}
    \vspace{-0.2in}
	\caption{Visualization of the feature distribution of the first and last stage of PoolFormer-S12 and RIFormer-S12.}
	\label{fig:feature}
    \vspace{-0.1in}
\end{figure}

\paragraph{Module imitation (MI) shifts the feature distribution of the RIFormer model to be closer to the teacher.} The effect of the module imitation is explicitly shown in Fig.~\ref{fig:feature}. It can be observed that PoolFomer-S12 and RIFormer-S12 show a clear difference in feature distribution of Stage 1 and Stage 4. After applying the proposed module imitation, the distribution of RIFormer-S12 are basically shifted toward that of the PoolFomer-S12, demonstrating its effect on helping student learn useful knowledge from the teacher. 

\vspace{-0.15in}
\paragraph{Module imitation helps showing larger Effective Receptive Field (ERF).} ERF~\cite{luo2016understanding} reflects how large an area of the image the trained model can respond to or capture information about how large an object. We follow~\cite{ding2022scaling, kim2021dead} to visualize the ERF via measuring the aggregated contributions of each pixel of the input to the central points of the output feature.
Since RIFormer removes all token mixers, it exhibits an expectedly much smaller ERF than PoolFormer, as shown in Fig.~\ref{fig:erf}. There is only a square of pixels emerging with red color in the whole region, much smaller than PoolFormer. However, surprisingly, we can observe that red color widely distributed in all locations after training with module imitation. It seems that although there's no explicit structural change, module imitation still help change the learned weights and show larger ERF.
\begin{table}[t]
    \begin{center}
        \small
        \begin{tabular}{lcccc}
            \hline
            Model & Type  & Throughput &  Top-1 (\%)	\\
            \hline
            PoolFormer-S12    &	None		& 4160.18 &	75.01		\\
            \hline
            PoolFormer-S9    &	Depth		& 5025.71 &	74.78		\\
            % \hline
            PoolFormer-XS12    &	Width		& 4780.28 &	75.11		\\
            
            \gr
            RIFormer-S12		&	TM		& 4899.80 &	\textbf{75.36}		\\
            \hline
        \end{tabular}
    \end{center}
    \vspace{-0.15in}
    \caption{Results of comparison with depth or width slimming.}
    \label{table-ab-2}
    \vspace{-0.1in}
\end{table}
\begin{table}[t]
    \begin{center}
        \small
        \begin{tabular}{lcccc}
            \hline
            Token Mixer & Teacher   &   Top-1 (\%)	\\
            \hline
            Affine (12 layers)  &	None	&	72.75		\\
            \hline
            \gr
            Affine (12 layers)	 &	RandFormer-S12~\cite{yu2022metaformerbase}		&	\textbf{75.62}		\\
            \gr
            Affine (12 layers)	 &	PoolFormer V2-S12~\cite{yu2022metaformerbase}		&	\textbf{75.87}		\\
            \hline
            Affine (18 layers)  &	None	&	75.01		\\
            \hline
            \gr
            Affine (18 layers)	 &	ConvFormer-S18~\cite{yu2022metaformerbase}		&	\textbf{77.53}		\\
            \gr
            Affine (18 layers)  &	CAFormer-S18~\cite{yu2022metaformerbase}	&	\textbf{77.26}		\\
            \hline
        \end{tabular}
    \end{center}
    \vspace{-0.22in}
    \caption{Results of generalization to other teachers.}
    \label{table-ab-3}
    \vspace{-0.24in}
\end{table}

\section{Limitations and Conclusion}
\label{sec:conclusion}
% \vspace{-0.5em}
This paper investigates removing token mixer of the basic building block in a vision backbone, motivated by their considerable latency cost. To keep the remaining structure still effective, we thoroughly revisits the training paradigm. We observe that appropriate optimization strategy can effectively help a token mixer-free model learn  useful knowledge from another model, boosting its performance and bridge the gap caused by incomplete structure. Limitations are that more vision tasks, including detection, deblurring, \etc are not discussed, and we will work on them in the future. 

\section*{Acknowledgement}
This paper is partly supported by the National Key R\&D Program of China No.2022ZD0161000, the General Research Fund of HK No.17200622, Shanghai Post-doctoral Excellence Program (No.2022235), the National Natural Science Foundation of China (Grant No.61991450) and the Shenzhen Science and Technology Program (JCYJ20220818101001004).

%%%%%%%%% REFERENCES
{\small
\bibliographystyle{ieee_fullname}
\bibliography{egbib}
}

\clearpage
\appendix

\section{Detailed hyper-parameters of Sec.4}
\label{sec:appendix.1}
We provide some experimental settings of the exploration roadmap of Sec.4 in the main paper. Generally, we use a RIFormer-S12 model in this section, which is trained and evaluated on ImageNet-1K for 120 epochs. We take AdamW~\cite{loshchilov2017decoupled, kingma2014adam} optimizer with batch size of 512 in all circumstances.
For distillation experiments in Sec.4.2 and Sec.4.3, GFNet-H-B~\cite{rao2021global} serves as the teacher with a logit distillation following~\cite{deit}. 

\section{Proof of Eq.4}
\label{sec:appendix.2}

Given $\mathrm{T}^{(a)}\in\mathbb{R}^{N\times C\times H\times W}$, ${\mathrm{T}^{\prime(a)}}\in\mathbb{R}^{N\times C\times H\times W}$ respectively as the input and output of an affine residual sub-block in Fig.2-(a) in the main paper. During the training time, we have:
% \vspace{-0.15in}
\begin{equation}
\begin{aligned}
\mathrm{T}^{\prime(a)} &= \mathrm{Affine}(\mathrm{LN}(\mathrm{T}^{(a)},\vect{\mu},\vect{\sigma},\vect{\gamma},\vect{\beta}),\boldsymbol{\mathbf{s}},\boldsymbol{\mathbf{t}}) -\mathrm{T}^{(a)}
\end{aligned}
\label{equ:supple_1}
\end{equation}
where $\mathrm{LN}$ is the LN layer, $\vect{\mu},\vect{\sigma},\vect{\gamma},\vect{\beta}$ as the mean, standard deviation and learned scaling factor and bias of the LN layer, $\mathrm{Affine}$ is the affine transformation, $\mathbf{s}\in\mathbb{R}^{C}$ and $\mathbf{t}\in\mathbb{R}^{C}$ are its learnable scaling and shift parameters. During the training time, we have:
\begin{equation}
\begin{aligned}
\mathrm{T}^{\prime(a)} &= \mathrm{LN}(\mathrm{T}^{(a)},\vect{\mu},\vect{\sigma},\vect{\gamma}^\prime,\vect{\beta}^\prime)
\end{aligned}
\label{equ:supple_2}
\end{equation}
According to the equivalence of the structural re-parameterization of the affine residual sub-block during training (Eq.~\ref{equ:supple_1}) and inference (Eq.~\ref{equ:supple_2}), for $\forall 1\leq n \leq N$, $\forall 1\leq i \leq C$, $\forall 1\leq h \leq H$, $\forall 1\leq w \leq W$, we have:
\begin{equation}
\begin{aligned}
& ((\mathrm{T}^{(a)}_{n,i,h,w} - \boldsymbol{\mathbf{\mu}}_{n,h,w})\frac{\boldsymbol{\mathbf{\gamma}}_i}{\boldsymbol{\mathbf{\sigma}}_{n,h,w}} + \boldsymbol{\mathbf{\beta}}_i)\boldsymbol{\mathbf{s}}_i + \boldsymbol{\mathbf{t}}_i\\ 
& ~~~~ - (\mathrm{T}^{(a)}_{n,i,h,w} - \boldsymbol{\mathbf{\mu}}_{n,h,w})\frac{\boldsymbol{\mathbf{\gamma}}_i}{\boldsymbol{\mathbf{\sigma}}_{n,h,w}} -\boldsymbol{\mathbf{\beta}}_i \\
& ~~~~ = (\mathrm{T}^{(a)}_{n,i,h,w} - \boldsymbol{\mathbf{\mu}}_{n,h,w})\frac{\boldsymbol{\mathbf{\gamma}}^\prime_i}{\boldsymbol{\mathbf{\sigma}}_{n,h,w}} + \boldsymbol{\mathbf{\beta}}^\prime_i
\end{aligned}
\label{equ:supple_3}
\end{equation}
Eq.~\ref{equ:supple_3} can be reformulated as:
\begin{equation}
\begin{aligned}
\vect{\gamma}^\prime_i = \vect{\gamma}_i(\boldsymbol{\mathbf{s}}_i-1),~~~ 
\vect{\beta}^\prime_i = \vect{\beta}_i(\boldsymbol{\mathbf{s}}_i-1)  +\boldsymbol{\mathbf{t}}_i, 
\end{aligned}
\label{equ:supple_4}
\end{equation}
Then Eq.4 in the main paper follows. 

\section{Code in PyTorch}
\label{sec:appendix.3}
\subsection{PyTorch-like code of our affine operator}
\label{sec:appendix.3.1}
We provide the PyTorch-like code of the affine transformation in Alg.~\ref{alg:affine} affiliated with the training-time model in the RIFormer block. The affine transformation can be implemented as a depth-wise convolution by specifying the kernel size as $1$ and the group number as the input channels. We follow~\cite{yu2022metaformer} to add a subtraction of the input during implementation due to the residual connection.

% The proof and PyTorch-like code of the affine transformation and re-parameterization process is shown in Sec.2 and Sec.3 of the appendix, respectively.
\begin{algorithm}[t]
\caption{Affine Transformation, PyTorch-like Code}
\label{alg:affine}
\definecolor{codeblue}{rgb}{0.25,0.5,0.5}
\definecolor{codekw}{rgb}{0.85, 0.18, 0.50}
\lstset{
  backgroundcolor=\color{white},
  basicstyle=\fontsize{7.5pt}{7.5pt}\ttfamily\selectfont,
  columns=fullflexible,
  breaklines=true,
  captionpos=b,
  commentstyle=\fontsize{7.5pt}{7.5pt}\color{codeblue},
  keywordstyle=\fontsize{7.5pt}{7.5pt}\color{codekw},
}
\begin{lstlisting}[language=python]
import torch.nn as nn

class Affine(nn.Module):
    def __init__(self, in_features):
        super().__init__()
        self.affine = nn.Conv2d(
            in_features, in_features, kernel_size=1, stride=1, padding=0, groups=in_features, bias=True)

    def forward(self, x):
        """
        [B, C, H, W] = x.shape
        Subtraction of the input itself is added 
        since the block already has a 
        residual connection.
        """
        return self.affine(x) - x
\end{lstlisting}
\end{algorithm}

\subsection{PyTorch-like code of the RIFormer block}
\label{sec:appendix.3.2}
We provide the PyTorch-like code of the RIFormer block with the strctural re-parameterization in Alg.~\ref{alg:rep}. 
\begin{algorithm*}[t]
\caption{RIFormer Block (dubbed as AffineFormerBlock), PyTorch-like Code}
\label{alg:rep}
\definecolor{codeblue}{rgb}{0.25,0.5,0.5}
\definecolor{codekw}{rgb}{0.85, 0.18, 0.50}
\lstset{
  backgroundcolor=\color{white},
  basicstyle=\fontsize{7.5pt}{7.5pt}\ttfamily\selectfont,
  columns=fullflexible,
  breaklines=true,
  captionpos=b,
  commentstyle=\fontsize{7.5pt}{7.5pt}\color{codeblue},
  keywordstyle=\fontsize{7.5pt}{7.5pt}\color{codekw},
}
\begin{lstlisting}[language=python]
import torch.nn as nn

class AffineFormerBlock(nn.Module):

    def __init__(self, dim, mlp_ratio=4., act_layer=nn.GELU, norm_layer=GroupNorm, 
                 drop=0., drop_path=0., 
                 use_layer_scale=True, layer_scale_init_value=1e-5, deploy=False):
        super().__init__()
        if deploy:
            self.norm_reparam = norm_layer(dim)
        else:
            self.norm1 = norm_layer(dim)
            self.token_mixer = Affine(in_features=dim)
        self.norm2 = norm_layer(dim)
        mlp_hidden_dim = int(dim * mlp_ratio)
        self.mlp = Mlp(in_features=dim, hidden_features=mlp_hidden_dim, 
                       act_layer=act_layer, drop=drop)

        # The following two techniques are useful to train deep AffineFormers.
        self.drop_path = DropPath(drop_path) if drop_path > 0. \
            else nn.Identity()
        self.use_layer_scale = use_layer_scale
        if use_layer_scale:
            self.layer_scale_1 = nn.Parameter(
                layer_scale_init_value * torch.ones((dim)), requires_grad=True)
            self.layer_scale_2 = nn.Parameter(
                layer_scale_init_value * torch.ones((dim)), requires_grad=True)
        self.norm_layer = norm_layer
        self.dim = dim
        self.deploy = deploy

    def forward(self, x):
        if hasattr(self, 'norm_reparam'):
            if self.use_layer_scale:
                x = x + self.drop_path(
                    self.layer_scale_1.unsqueeze(-1).unsqueeze(-1)
                    * self.norm_reparam(x))
                x = x + self.drop_path(
                    self.layer_scale_2.unsqueeze(-1).unsqueeze(-1)
                    * self.mlp(self.norm2(x)))
            else:
                x = x + self.drop_path(self.norm_reparam(x))
                x = x + self.drop_path(self.mlp(self.norm2(x)))           
        else:
            if self.use_layer_scale:
                x = x + self.drop_path(
                    self.layer_scale_1.unsqueeze(-1).unsqueeze(-1)
                    * self.token_mixer(self.norm1(x)))
                x = x + self.drop_path(
                    self.layer_scale_2.unsqueeze(-1).unsqueeze(-1)
                    * self.mlp(self.norm2(x)))
            else:
                x = x + self.drop_path(self.token_mixer(self.norm1(x)))
                x = x + self.drop_path(self.mlp(self.norm2(x)))
        return x

    def fuse_affine(self, norm, token_mixer):
        gamma_affn = token_mixer.affine.weight.reshape(-1)
        gamma_affn = gamma_affn - torch.ones_like(gamma_affn)
        beta_affn = token_mixer.affine.bias
        gamma_ln = norm.weight
        beta_ln = norm.bias
        print('gamma_affn:', gamma_affn.shape)
        print('beta_affn:', beta_affn.shape)
        print('gamma_ln:', gamma_ln.shape)
        print('beta_ln:', beta_ln.shape)
        return (gamma_ln * gamma_affn), (beta_ln * gamma_affn + beta_affn)

    def get_equivalent_scale_bias(self):
        eq_s, eq_b = self.fuse_affine(self.norm1, self.token_mixer)
        return eq_s, eq_b
        
    def switch_to_deploy(self):
        if self.deploy:
            return
        eq_s, eq_b = self.get_equivalent_scale_bias()
        self.norm_reparam = self.norm_layer(self.dim)
        self.norm_reparam.weight.data = eq_s
        self.norm_reparam.bias.data = eq_b
        self.__delattr__('norm1')
        if hasattr(self, 'token_mixer'):
            self.__delattr__('token_mixer')
        self.deploy = True
\end{lstlisting}
\end{algorithm*}

\section{Detailed hyper-parameters on ablations}
\label{sec:appendix.4}
We provide the experimental settings of ablation studies of Sec.5 in the main paper. 

For the $74.90\%$ and $75.36\%$ top-1 accuracy experiments in Tab.7 in the main paper, GFNet-H-B~\cite{rao2021global} serves as the teacher with a logit distillation, and PoolFormer-S12~\cite{yu2022metaformer} serves as the teacher for adopting the proposed module imitation strategy. For $(\mathcal{L}_{in}^\prime, \mathcal{L}_{out})$ and $\mathcal{L}_{rel}$ in Eq.9 in the main paper, the number of epochs of using them are 80 and 20, respectively. The differences are as follows. In the $74.90\%$ experiment, we set $\lambda_2=\lambda_3=0$ in Eq.9 to obtain a hidden state feature distillation. In the $75.36\%$ experiment, we choose $\lambda_2$ and $\lambda_3$ as in Tab.~\ref{tab:hyperparameter-imagenet}, and initialize the weights of RIFormer (except the affine operator) with the corresponding teacher network, as presented in Sec.4.4 of the main paper.

For Tab.9 in the main paper, we adopt the architectural modifications in~\cite{yu2022metaformerbase} to our RIFormer, and construct two models with 12 and 18 layers as our student model, respectively. For distillation experiments in Tab.9, GFNet-H-B~\cite{rao2021global} serves as the teacher with a logit distillation.  RandFormer-S12~\cite{yu2022metaformerbase}, PoolFormer V2-S12~\cite{yu2022metaformerbase}, ConvFormer-S18~\cite{yu2022metaformerbase}, CAFormer-S18~\cite{yu2022metaformerbase} serve as teacher networks for module imitation. We train and evaluate on ImageNet-1K for 130 epochs (with 10 patient epochs).

\section{Detailed hyper-parameters of ImageNet-1K}
\label{sec:appendix.5}
We provide the experimental settings of ImageNet-1K classification of Sec.5 in the main paper in Tab.~\ref{tab:hyperparameter-imagenet}. The hyper-parameters generally follow~\cite{yu2022metaformer}. We use AdamW~\cite{loshchilov2017decoupled} optimizer with a batch size of $1024$ and weight decay of $0.05$, and learning rate $\mathrm{lr} = 1e^{-3} \cdot \mathrm{batch\ size} / 1024$. Stochastic Depth\cite{stochastic_depth}, Label Smoothing\cite{label_smoothing} and Layer Scale\cite{touvron2021going} are also adopted to regularize the networks. For the RIFormer result with $224^2$ input resolution in Tab.6 in the main paper, GFNet-H-B~\cite{rao2021global} serves as the teacher with a logit distillation, and a PoolFormer~\cite{yu2022metaformer} with same parameter size serves as the teacher for adopting the proposed module imitation strategy. For the $384^2$ RIFormer finetuning results, we use ConvFormer~\cite{yu2022metaformerbase} as the teacher with a logit distillation and do not perform module imitation in the step. For the throughput measurement, 
we take a batch size of 2048 at $224^2$ resolution (1024 for $384^2$ resolution) with one 80GB A100 GPU and calculate the average throughput over 30 runs to inference that batch. The process is repeated for three times and the medium are treated as the statistical throughput. 

\begin{table*}[htbp]
\centering
\begin{tabular}{@{}l|ccccc@{}}
\toprule
 & \multicolumn{5}{c}{RIFormer} \\ 
 & S12 & S24 & S36 & M36 & M48 \\
\midrule
Peak drop rate of stoch. depth $d_r$ & 0.1 & 0.1 & 0.1 & 0.1 & 0.1 \\
LayerScale initialization $\epsilon$ & $10^{-5}$ & $10^{-5}$ & $10^{-6}$ & $10^{-6}$ & $10^{-6}$ \\
$\lambda_1\times\mathrm{batch\_size}$ in Eq.9 & $0.0001$ & $0.0003$ & $0.0001$ & $0.0001$ & $0.0001$ \\
$\lambda_2\times\mathrm{batch\_size}$ in Eq.9 & $0.001$ & $0.005$ & $0.001$ & $0.001$ & $0.001$\\
$\lambda_3\times\mathrm{batch\_size}$ in Eq.9 & $1.0$ & $4.0$ & $1.0$ & $1.0$ & $1.0$\\
\hline
Data augmentation & \multicolumn{5}{c}{AutoAugment} \\
Repeated Augmentation & \multicolumn{5}{c}{off} \\
Input resolution & \multicolumn{5}{c}{224} \\
Epochs & \multicolumn{5}{c}{600} \\
Number of epochs of using $(\mathcal{L}_{in}^\prime, \mathcal{L}_{out})$ & \multicolumn{5}{c}{400} \\
Number of epochs of using $\mathcal{L}_{rel}$ & \multicolumn{5}{c}{100} \\
Warmup epochs & \multicolumn{5}{c}{5} \\
Hidden dropout & \multicolumn{5}{c}{0} \\
GeLU dropout & \multicolumn{5}{c}{0} \\
Classification dropout & \multicolumn{5}{c}{0} \\
Random erasing prob & \multicolumn{5}{c}{0.25} \\
EMA decay & \multicolumn{5}{c}{0} \\
Cutmix $\alpha$ & \multicolumn{5}{c}{1.0} \\
Mixup $\alpha$ & \multicolumn{5}{c}{0.8} \\
Cutmix-Mixup switch prob & \multicolumn{5}{c}{0.5} \\
Label smoothing & \multicolumn{5}{c}{0.1} \\
\tabincell{l}{Relation between peak learning \\ \qquad rate and batch size} & \multicolumn{5}{c}{$\mathrm{lr} = \frac{\mathrm{batch\_size}}{1024}\times 10^{-3}$} \\
Batch size used in the paper & $1024$ & $1024$ & $1024$ & $1024$ & $512$\\
Peak learning rate used in the paper & \multicolumn{5}{c}{$1 \times 10^{-3}$} \\
Learning rate decay & \multicolumn{5}{c}{cosine} \\
Optimizer & \multicolumn{5}{c}{AdamW} \\
Adam $\epsilon$ & \multicolumn{5}{c}{1e-8} \\
Adam $(\beta_1, \beta_2)$ & \multicolumn{5}{c}{(0.9, 0.999)} \\
Weight decay & \multicolumn{5}{c}{0.05} \\
Gradient clipping & \multicolumn{5}{c}{None} \\
\bottomrule
\end{tabular}
\vspace{-2mm}
\caption{\textbf{Hyper-parameters for image classification on ImageNet-1K}
\label{tab:hyperparameter-imagenet}
}
\end{table*}

\section{Visualization of the learned coefficients}
\label{sec:appendix.6}
\begin{figure*}[t]
	\centering
	\includegraphics[width=1.0\linewidth]{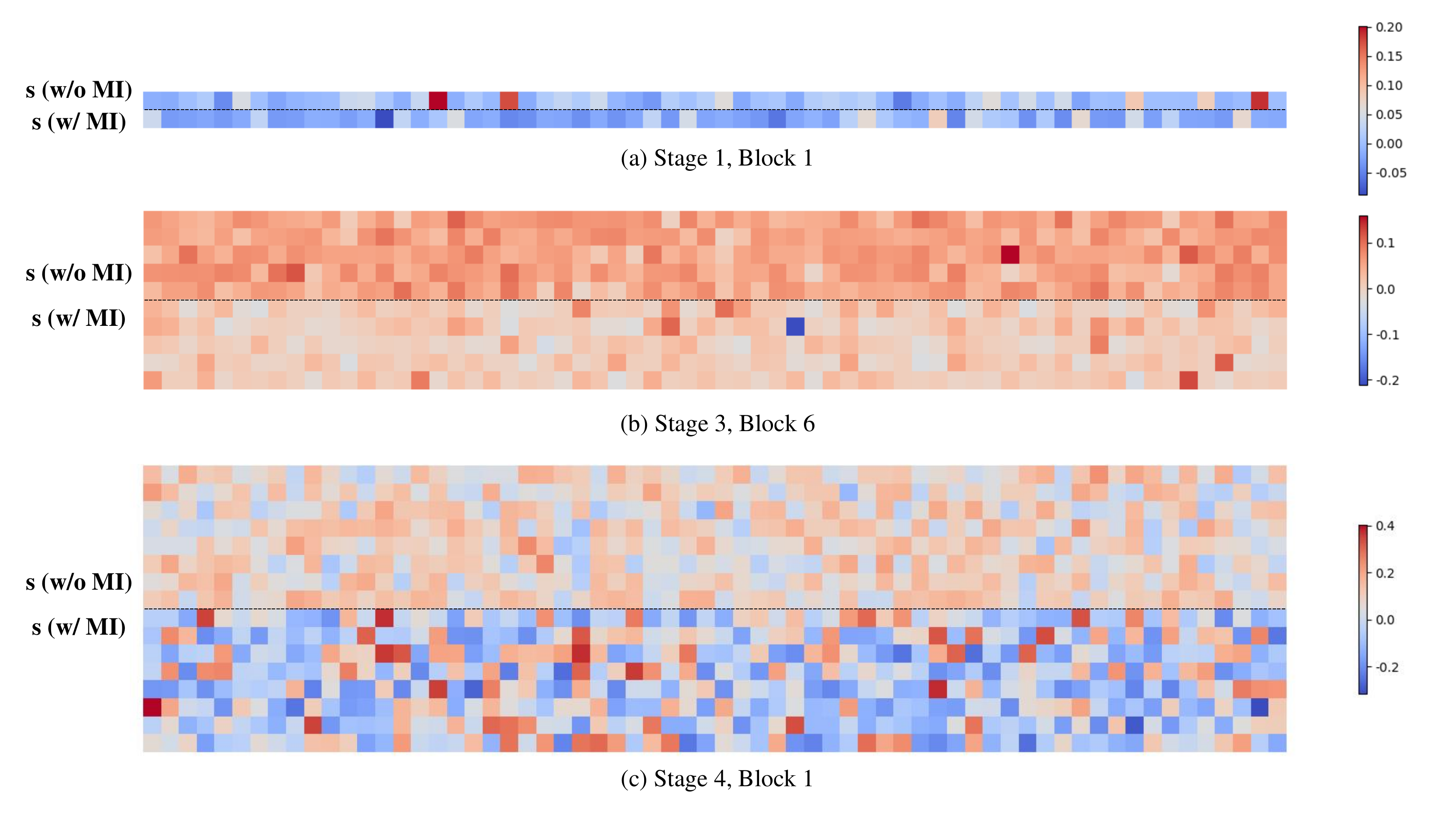}
	\caption{The heatmap of the learned coefficients of the affine transformation in (a) Stage 1, Block 1, (b) Stage 3, Block 6, (c) Stage 4, Block 1, respectively. The values of the learned coefficients are given different colors for positive and negative number.}
	\label{fig:weight}
\end{figure*}

To further evaluate the effect of the proposed module imitation algorithm, we visualize the learned coefficients of the weights (denoted as $\mathbf{s}$) of the affine operator with (above the black dotted line) or without (below the black dotted line) the module imitation technique. Specifically, we provide the learned affine weights of a shallow block (Stage 1, Block 1), an intermediate block (Stage 3, Block 6), and a deep block (Stage 4, Block 1). As shown in Fig.~\ref{fig:weight}, the affine weights trained using module imitation show differences with those of trained without such technique. Take Fig.~\ref{fig:weight}-(c) as an instance. The affine weights without using module imitation are relatively more consistent and appear to show more positive values. As a comparison, module imitation help the affine operator learn more diverse and negative weights, which may useful for the expressiveness of our RIFormer. Similarly in Fig.~\ref{fig:weight}-(b), the affine weights using module imitation have more moderate amplitude, compared to a higher amplitude of without the method.

\section{Visualization of the activation parts}
\label{sec:appendix.7}
Following~\cite{yu2022metaformer}, we provide qualitative results of four different pre-trained backbones obtained by Grad-CAM~\cite{selvaraju2017grad}, respectively RSB-ResNet50~\cite{he2016deep, wightman2021resnet}, DeiT-S~\cite{deit}, PoolFormer-S24~\cite{yu2022metaformer} and our RIFormer-S24. As observed in~\cite{yu2022metaformer}, the activation parts in the map of a transformer model are scattered, while that of a convnet are more aggregated. Interestingly, two additional observation can be made. Firstly, it seems that RIFormer trained with the proposed module imitation algorithm combines the characteristics of both convnet and transformer. We deem the reason might be that RIFormer has the same general architecture as transformer, but without any attention (\ie, token mixer), and thus it is essentially a convnet. Secondly, the activation parts in the RIFormer map of show similar characteristics of that in PoolFormer, which may be due to the inductive bias implicitly incorporated from the teacher model via the knowledge distillation process.
\begin{figure*}[t]
    % \vspace{-3mm}
    \centering
    \begin{subfigure}[b]{0.19\textwidth}
        \centering
        \includegraphics[width=1\textwidth]{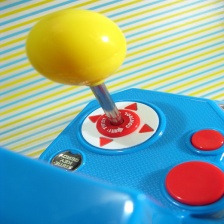}
        \includegraphics[width=1\textwidth]{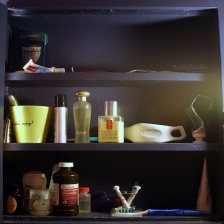}
        \includegraphics[width=1\textwidth]{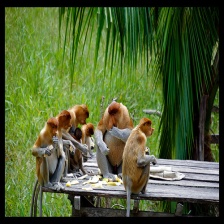}
        \includegraphics[width=1\textwidth]{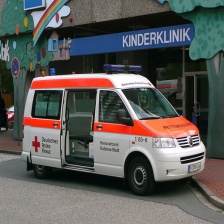}
    \end{subfigure}    
    \begin{subfigure}[b]{0.19\textwidth}
        \centering
        \includegraphics[width=1\textwidth]{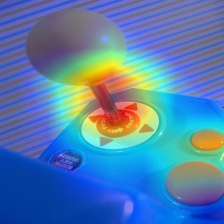}
        \includegraphics[width=1\textwidth]{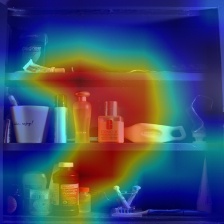}
        \includegraphics[width=1\textwidth]{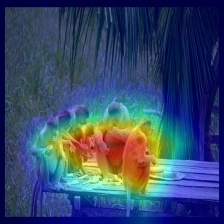}
        \includegraphics[width=1\textwidth]{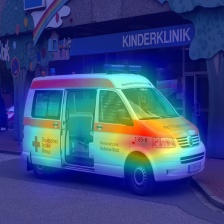}
    \end{subfigure}  
    \begin{subfigure}[b]{0.19\textwidth}
        \centering
        \includegraphics[width=1\textwidth]{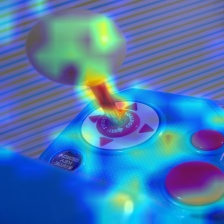}
        \includegraphics[width=1\textwidth]{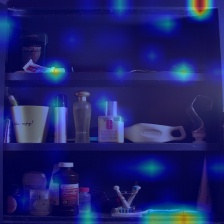}
        \includegraphics[width=1\textwidth]{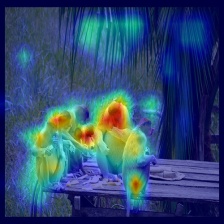}
        \includegraphics[width=1\textwidth]{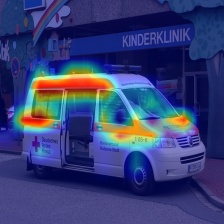}
    \end{subfigure}  
    \begin{subfigure}[b]{0.19\textwidth}
        \centering
        \includegraphics[width=1\textwidth]{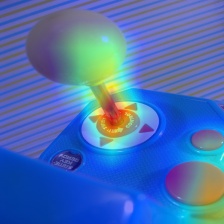}
        \includegraphics[width=1\textwidth]{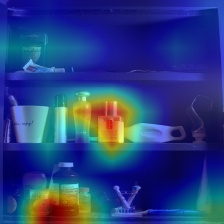}
        \includegraphics[width=1\textwidth]{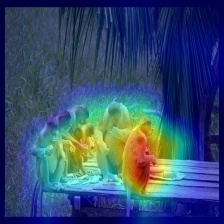}
        \includegraphics[width=1\textwidth]{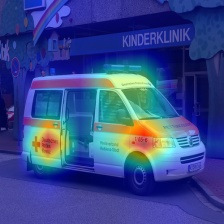}
    \end{subfigure} 
    \begin{subfigure}[b]{0.19\textwidth}
        \centering
        \includegraphics[width=1\textwidth]{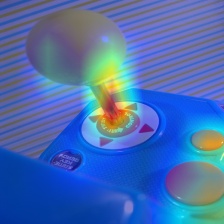}
        \includegraphics[width=1\textwidth]{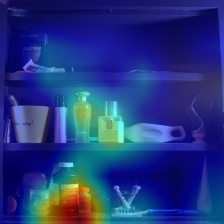}
        \includegraphics[width=1\textwidth]{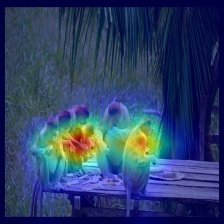}
        \includegraphics[width=1\textwidth]{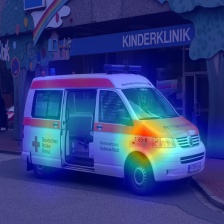}
    \end{subfigure}
    % \vspace{-7mm}
    \begin{center}
    	 ~~~~~~~Input\qquad \qquad \quad RSB-ResNet-50~\cite{wightman2021resnet} \qquad ~~ DeiT-small \cite{deit}~\qquad ~~ PoolFormer-S24~\cite{yu2022metaformer} \qquad ~~ RIFormer-S24~
    \end{center}  
    \caption{
        \label{grad_cam} Grad-CAM \cite{selvaraju2017grad} activation maps of four different pre-trained backbones on ImageNet-1K. We sample 4 images to visualize from the validation set.
    }
\end{figure*}

\end{document}